\documentclass[11pt]{article}
\usepackage{coling2020}
\usepackage{times}
\usepackage{url}
\usepackage{latexsym}
\usepackage{tabularx}
\usepackage{booktabs}
\usepackage{todonotes}
\usepackage{multicol}
\usepackage{multirow}

\usepackage{comment}
\usepackage{caption}
\usepackage{makecell}
\usepackage{graphicx}
\usepackage{colortbl}
\usepackage{microtype}

\DeclareCaptionLabelFormat{AppendixTables}{A.#2}

\title{HateBERT: Retraining BERT for Abusive Language Detection in English}

\author{Tommaso Caselli$^{\clubsuit}$, Valerio Basile$^{\diamond}$, Jelena Mitrovi\'{c}$^{\ddagger}$, Michael Granitzer$^{\ddagger}$ \\
$^{\clubsuit}$University of Groningen, $\diamond$University of Turin, $\ddagger$University of Passau \\
         Groningen The Netherlands, Turin Italy, Passau Germany  \\
\tt $\diamond$\{valerio.basile\}@unito.it, 
 $\clubsuit$t.caselli@rug.nl \\ \tt $\ddagger${jelena.mitrovic|michael.granitzer\}@uni-passau.de}\\
}

\date{}

\begin{document}
\maketitle
\begin{abstract}
  In this paper, we introduce HateBERT, a re-trained BERT model for abusive language detection in English. The model was trained on RAL-E, a large-scale dataset of Reddit comments in English from communities banned for being offensive, abusive, or hateful that we have collected and made available to the public. We present the results of a detailed comparison between a general pre-trained language model and the abuse-inclined version obtained by retraining with posts from the banned communities on three English datasets for offensive, abusive language and hate speech detection tasks. In all datasets, HateBERT outperforms the corresponding general BERT model. We also discuss a battery of experiments comparing the portability of the generic pre-trained language model and its corresponding abusive language-inclined counterpart across the datasets, indicating that portability is affected by compatibility of the annotated phenomena.
\end{abstract}

\section{Introduction}

The popularity of social media and micro-blogging platforms is still having undisclosed effects in our life as a result of an increased connectivity among people. The potential benefits are overshadowed by numerous expressions of abusive language phenomena. This contribution focuses on finding solutions to overcome that problem.

The development of systems for the automatic identification of abusive language phenomena has followed a common trend in NLP: feature-based linear classifiers~\cite{waseem-hovy:2016:N16-2,ribeiro2018characterizing,ibrohim2019multi}, neural network architectures (e.g., CNN or Bi-LSTM)~\cite{kshirsagar-etal-2018-predictive,mishra-etal-2018-neural,mitrovic-etal-2019-nlpup,sigurbergsson2020offensive}, and, finally, fine-tuning pre-trained language models, e.g., BERT, RoBERTa, a.o.,~\cite{liu-etal-2019-nuli,swamy-etal-2019-studying}. Results vary both across datasets and architectures, with linear classifiers qualifying as very competitive, if not better, when compared to neural networks. On the other hand, systems based on pre-trained language models have reached new state-of-the-art results. One issue with these pre-trained models is that the training language variety makes them well suited for general-purpose language understanding tasks, and it highlights their limits with more domain-specific language varieties. To address this, there is a growing interest in generating domain-specific BERT-like pre-trained language models, such as AlBERTo~\cite{DBLP:conf/aiia/PolignanoBGS19} or TweetEval~\cite{barbieri-etal-2020-tweeteval} for Twitter, BioBERT for the biomedical domain in English~\cite{biobert}, FinBERT for the financial domain in English~\cite{finbert}, and LEGAL-BERT for the legal domain in English~\cite{chalkidis-etal-2020-legal}. We introduce HateBERT, a pre-trained BERT model for abusive language phenomena in social media in English.

\begin{figure}[!thb]
    \centering
    \includegraphics[width=.45\textwidth]{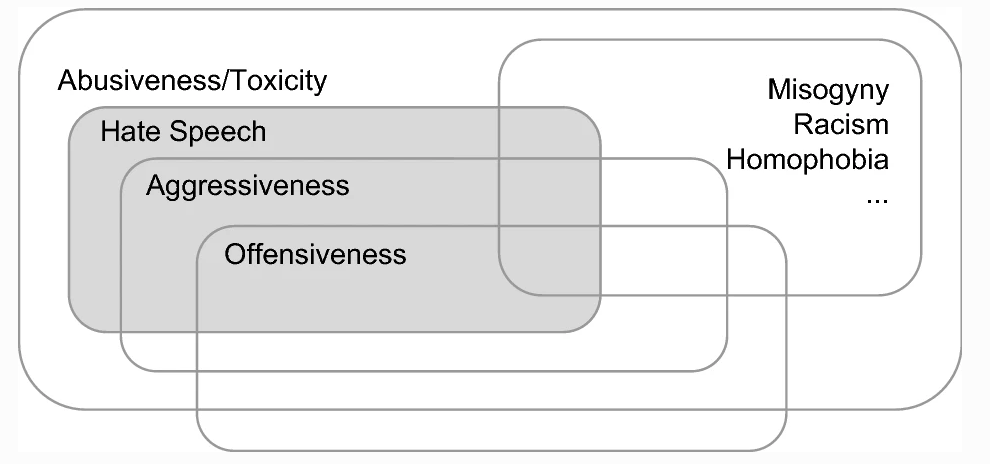}
    \caption{Abusive language phenomena and their relationships (source \newcite{Poletto2020}).}
    \label{fig:phenomena}
\end{figure}

Abusive language phenomena fall along a wide spectrum including, a.o., microaggression, stereotyping, offense, abuse, hate speech, threats, and doxxing ~\cite{jurgens-etal-2019-just}. Current approaches have focus on a limited range, namely offensive language, abusive language, and hate speech. The connections among these phenomena have only superficially been accounted for, resulting in a fragmented picture, with a variety of definitions, and (in)compatible annotations~\cite{waseem2017understanding}. ~\newcite{Poletto2020} introduce a graphical visualisation (Figure~\ref{fig:phenomena}) of the connections among abusive language phenomena according to the definitions in previous work~\cite{waseem-hovy:2016:N16-2,fortuna2018survey,malmasi2018challenges,basile-etal-2019-semeval,zampieri-etal-2019-semeval}.  

When it comes to offensive language, abusive language, and hate speech, the distinguishing factor is their level of specificity. This makes offensive language the most generic form of abusive language phenomena and hate speech the most specific, with abusive language being somewhere in the middle. Such differences are a major issue for the study of portability of models.  Previous work~\cite{karan-snajder-2018-cross,benk2019data,pamungkas2019cross,rizoiu2019transfer} has addressed this task by conflating portability with generalizability, forcing datasets with different phenomena into homogenous annotations by collapsing labels into (binary) macro-categories. In our portability experiments, we show that the behavior of HateBERT can be explained by accounting for these difference in specificity across the abusive language phenomena. 

Our key contributions are: (i.) additional evidence that further pre-training is a viable strategy to obtain domain-specific or language variety-oriented models in a fast and cheap way; (ii.) the release of HateBERT, a pre-trained BERT for abusive language phenomena, intended to boost research in this area; (iii.) the release of a large-scale dataset of social media posts in English from communities banned for being offensive, abusive, or hateful.

\section{HateBERT: Re-training BERT with Abusive Online Communities}

While transformer-based pre-trained language models, such as BERT, achieve good performance on numerous NLP tasks, when applied to less standard language varieties, such as social media data, results may fluctuate a lot. For instance, by comparing different fine-tuned BERT models on the OffensEval 2019 dataset~\cite{zampieri-etal-2019-semeval}, it appears that the key factor in boosting the performance is the quality of the pre-processing step~\cite{liu-etal-2019-nuli,swamy-etal-2019-studying}, rather than other aspects such as learning rate or training time. 

Further pre-training of a BERT-like model is becoming more and more popular as a competitive, effective, and fast solution to adapt pre-trained language models to new language varieties or domains~\cite{barbieri-etal-2020-tweeteval,biobert,finbert,chalkidis-etal-2020-legal}, especially in cases where raw data are not scarce to generate a BERT-like model from scratch~\cite{gururangan-etal-2020-dont}. This is the case of abusive language phenomena. However, for these phenomena an additional predicament with respect to previous work is that the options for suitable and representative collections of data are very limited. Directly scraping messages containing profanities would not be the best option as lots of potentially useful data may be missed.~\newcite{essem} have used tweets about controversial topics to generate offensive-loaded embeddings, but their approach presents some limits. On the other hand,~\newcite{ijcol} have shown the effectiveness of using messages from potentially abusive-oriented on-line communities to generate so-called \textit{hate embeddings}. We follow this latter approach by using messages from banned communities in Reddit. 

\paragraph{RAL-E: the Reddit Abusive Language English dataset} Reddit is a popular social media outlet where users share and discuss content. The website is organized into 
user-created and user-moderated communities known as \textit{subreddits}, being \textit{de facto} on-line communities. In 2015, Reddit strengthened its content policies and banned several subreddits~\cite{chandrasekharan}. We retrieved a large list of banned communities in English from different sources including official posts by the Reddit administrators and Wikipedia pages.\footnote{\url{https://en.wikipedia.org/wiki/Controversial_Reddit_communities}} We then selected only communities that were banned for being deemed to host or promote offensive, abusive, and/or hateful content (e.g., expressing harassment, bullying, inciting/promoting violence, inciting/promoting hate). 
We collected the posts from these communities by crawling a publicly available collection of Reddit Comments from December 2005 to March 2017.\footnote{\url{https://www.reddit.com/r/datasets/comments/3bxlg7/i_have_every_publicly_available_reddit_comment/}} For each post, we kept only the text and the name of the community. 
The resulting collection comprises 1,492,740 messages from a period between January 2012 and June 2015, for a total of 43,820,621 tokens. The list of selected communities with the number of retrieved messages per community is reported in Table A.1 in Appendix A.
 
\paragraph{Creating HateBERT} From the RAL-E dataset, we used 1,478,348 messages (for a total of 43,379,350 tokens) to re-train the English BERT \texttt{base-uncased} model\footnote{We used the pre-trained model available via the huggingface Transformers library - \url{https://github.com/huggingface/transformers} 
} by applying the Masked Language Model (MLM) objective. The remaining 14,932 messages (441,271 tokens) have been used as test set. We retrained for 100 epochs (almost 2 million steps) in batches of 64 samples, including up to 512 sentencepiece tokens. We used Adam with learning rate \texttt{5e-5}. We trained using the huggingface code\footnote{\url{https://github.com/huggingface/transformers/tree/master/src/transformers}} on one Nvidia V100 GPU. The result is a shifted BERT model, HateBERT \texttt{base-uncased}, along two dimensions: (i.) language variety (i.e. social media); and (ii.) polarity (i.e., offense-, abuse-, and hate-oriented model).

\section{Experiments and Results}
\label{sec:experiments}

To verify the validity of HateBERT as being more suitable than a general one, i.e. BERT, for detecting offensive and abusive language phenomena, we run a set of experiments on three English datasets.

\paragraph{OffensEval 2019}~\cite{zampieri-etal-2019-semeval} This dataset was distributed in the context of the SemEval 2019: Task 6 evaluation exercise.\footnote{\url{https://competitions.codalab.org/competitions/20011}} The dataset contains 14,100 tweets annotated for \textbf{offensive} language. According to the task definition, a message is labelled as offensive if ``it contains any form of non-acceptable language (profanity) or a targeted offense, which can be veiled or direct.''~\cite[pg. 76]{zampieri-etal-2019-semeval}. 
The dataset is split into training and test, with 13,240 messages in training and 860 in test. The positive class (i.e. messages labeled as offensive) are 4,400 in training and 240 in test. No development data is provided.

\paragraph{AbusEval}~\cite{caselli-EtAl:2020:LREC} This dataset has been obtained by adding a layer of \textbf{abusive} language annotation to OffensEval 2019. Abusive language is defined as a specific case of offensive language, namely ``hurtful language that a speaker uses to insult or offend another individual or a group of individuals based on their personal qualities, appearance, social status, opinions, statements, or actions.''~\cite[pg. 6197]{caselli-EtAl:2020:LREC}. The main difference with respect to offensive language is the exclusion of isolated profanities or untargeted messages from the positive class. The size of the dataset is the same as OffensEval 2019, i.e., 14,100 tweets, as well as that of the training and test splits (13,240 and 860 messages, respectively). The differences concern the distribution of the positive class 
which results in 2,749 in training and 178 in test.

\paragraph{HatEval}~\cite{basile-etal-2019-semeval} This dataset was distributed for the SemEval 2019: Task 5 evaluation exercise.\footnote{\url{https://competitions.codalab.org/competitions/19935}} The English portion of the dataset contains 13,000 tweets annotated for \textbf{hate speech} against migrants and women. The authors 
define hate speech as ``any communication that disparages a person or a group on the basis of some characteristic such as race, color, ethnicity, gender, sexual orientation, nationality, religion, or other characteristics.''~\cite[pg. 54]{basile-etal-2019-semeval}. 
Only hateful messages targeting migrants and women belong to the positive class, leaving any other message (including offensive or abusive against other targets) to the negative class. The training set is composed of 10,000 messages and the test contains 3,000. Both training and test contain an equal amount of messages with respect to the targets, i.e., 5,000 each in training and 1,500 each in test. This does not hold for the distribution of the positive class, 
where 4,165 messages are present in the training and 1,252 in the test set.
\vspace{1em}

\noindent
All datasets are imbalanced between positive and negative classes and they target phenomena that vary along the specificity dimension. 
This allows us to evaluate both the robusteness and the portability of HateBERT.

We applied the same pre-processing steps and hyperparameters when fine-tuning both the generic BERT and HateBERT. Pre-processing steps and hyperparameters (Table A.2) are more closely detailed in the Appendix B. 
Table~\ref{tab:indomain} illustrates the results on each dataset (in-dataset evaluation), while Table~\ref{tab:crossdomain} reports on the portability experiments (cross-dataset evaluation). The same evaluation metric from the original tasks, or paper, is applied, that is macro-averaged F1 of the positive and negative classes.

\begin{table}[!th]
\centering
\small
\setlength{\tabcolsep}{1.0pt}
\begin{tabular}{llrr}
\toprule
\bf Dataset & \bf Model & \bf Macro F1 & \bf Pos. class - F1  \\ \midrule
\multirow{3}{*}{\makecell{OffensEval \\ 2019}} & BERT & .803$\pm$.006 & .715$\pm$.009 \\
& HateBERT & \bf .809$\pm$.008 & \bf .723$\pm$.012 \\ 
& \textit{Best} & .829 & .599 \\ \midrule
\multirow{3}{*}{AbusEval} & BERT & .727$\pm$.008 & .552$\pm$.012 \\
& HateBERT & \bf .765$\pm$.006 & \bf .623$\pm$.010  \\
& \newcite{caselli-EtAl:2020:LREC} & .716$\pm$.034 & .531 \\\midrule
\multirow{3}{*}{HatEval} & BERT & .480$\pm$.008 & .633$\pm$.002 \\
& HateBERT & \bf .516$\pm$.007 & \bf .645$\pm$.001 \\
& \textit{Best} & .651 & -- \\
\end{tabular}
\caption{BERT \textit{vs}. HateBERT: in-dataset. Best scores in bold. For BERT and HateBERT we report the average from 5 runs and its standard deviations. \textit{Best} corresponds to the best systems in the original shared tasks. \newcite{caselli-EtAl:2020:LREC} is the most recent result for AbusEval.}
\label{tab:indomain}
\end{table}

\begin{table}[ht]
\centering
\setlength{\tabcolsep}{1.0pt}
\small
\begin{tabular}{llrrr}
\toprule
\bf Train & \bf Model & \bf \makecell{OffensEval \\ 2019} & \bf AbusEval & \bf HatEval \\ \midrule
\multirow{2}{*}{\makecell{OffensEval \\ 2019}} & BERT & -- & .726 & .545 \\
& HateBERT & .-- & \underline{.750} & \underline{.547} \\ \midrule
\multirow{2}{*}{AbusEval} & BERT & .710 & -- & .611 \\ 
& HateBERT & \underline{.713} & -- & \underline{.624} \\ \midrule
\multirow{2}{*}{HatEval} & BERT & \underline{.572} & \underline{.590}  & -- \\
& HateBERT & .543 & .555 & -- \\
\end{tabular}
\caption{BERT \textit{vs}. HateBERT: Portability. Columns show the dataset used for testing. Best scores per training/test combination are underlined.}
\label{tab:crossdomain}
\end{table}

The in-domain results confirm the validity of the re-training approach to generate better models for detection of abusive language phenomena. On every dataset, HateBERT largely outperforms the corresponding general BERT model. A detailed analysis of the results per class show that the improvements, in all datasets, affect both the positive and the negative classes, suggesting that HateBERT is more robust. Interestingly, the use of data from a different social media platform does not harm the fine-tuning stage of the retrained model, opening up possibilities of cross-fertilisation studies across social media platforms. HateBERT beats the state-of-the-art results only in one case, namely for AbusEval, achieving competitive results on OffensEval and HatEval.\footnote{We did not manage to replicate the results by~\newcite{liu-etal-2019-nuli}, who qualified as the best system at OffensEval 2019.}  

The portability experiments were run using the best model for each of the in-dataset experiments. Besides a general drop in performance when compared to the corresponding in-dataset scores, our results show that HateBERT ensures better portability than a generic BERT model, especially when going from generic abusive language phenomena (i.e., offensive language) towards more specific one (i.e., abusive language or hate speech). This behaviour was expected and provides empirical evidence to the differences across the annotated phenomena. We also claim that HateBERT consistently obtains better representations of the targeted phenomena. This is evident when looking at the differences in False Positives and False Negatives for the positive class, measured by means of Precision and Recall, respectively. As illustrated in Table~\ref{tab:cross_p_r}, HateBERT always obtains a higher Precision score than BERT when fine-tuned on a generic abusive phenomenon and applied to more specific ones, at a very low cost for Recall. The unexpected higher Precision of HateBERT fine-tuned on AbusEval and tested on OffensEval 2019 (i.e., from specific to generic) is due to the datasets sharing same data distribution. Indeed, the results of the same model against HatEval support our analysis.  


\begin{table}[t]
\centering
\setlength{\tabcolsep}{1.0pt}
\small
\begin{tabular}{llrrrrrr}
\toprule
\bf Train & \bf Model & \multicolumn{2}{r}{\bf\makecell{\bf OffensEval \\ 2019}} & \multicolumn{2}{r}{\bf AbusEval} & \multicolumn{2}{r}{\bf HatEval} \\ \midrule
& & \bf P & \bf  R & \bf  P &\bf  R & \bf P & \bf R \\
\midrule
\multirow{2}{*}{\makecell{OffensEval \\ 2019}} & BERT & -- & -- & .510 & \underline{.685} & .479 & \underline{.771}  \\
& HateBERT & -- & -- & \underline{.553} & .696 & \underline{.480} & .767 \\ \midrule
\multirow{2}{*}{AbusEval} & BERT & .776 & \underline{.420} & -- & -- & .545 & \underline{.571}  \\ 
& HateBERT & \underline{.836} & .404 & -- & -- & \underline{.565} & .567 \\ \midrule
\multirow{2}{*}{HatEval} & BERT & \underline{.540} & \underline{.220}  & \underline{.438} & \underline{.241} & -- & -- \\
& HateBERT & .473 & .183 & .365 & .191 & -- & --\\
\end{tabular}
\caption{BERT \textit{vs}. HateBERT: Portability - Precision and Recall for the positive class. Rows show the dataset used to train the model and columns the dataset used for testing. Best scores are underlined.}
\label{tab:cross_p_r}
\end{table}

\section{Conclusion and Future Directions}

This contribution introduces HateBERT \texttt{base uncased},\footnote{HateBERT, the fine-tuned model, and the RAL-E dataset are available at \url{https://osf.io/tbd58/?view_only=cb79b3228d4248ddb875eb1803525ad8}} a pre-trained language model for abusive language phenomena in English. We confirm that further pre-training is an effective and cheap strategy to port pre-trained language models to other language varieties. 
The in-dataset evaluation shows that HateBERT consistently outperforms a generic BERT across 
different abusive language phenomena, such as offensive language (OffensEval 2019), abusive language (AbusEval), and hate speech (HatEval). The cross-dataset experiments show that HateBERT obtains robust representations of each abusive language phenomenon against which it has been fine-tuned. In particular, the cross-dataset experiments have provided (i.) further empirical evidence on the relationship among three abusive language phenomena along the dimension of specificity; (ii.) empirical support to the validity of the annotated data; (iii.) a principled explanation for the different performances of HateBERT and BERT.

Future work will focus on two directions: (i.) investigating to what extent the embedding representations of HateBERT are actually different from a general BERT pre-trained model, and (ii.) testing the generalizability of HateBERT.

\section*{Acknowledgements}
\begin{figure}[htbp]
   \includegraphics[width=6cm]{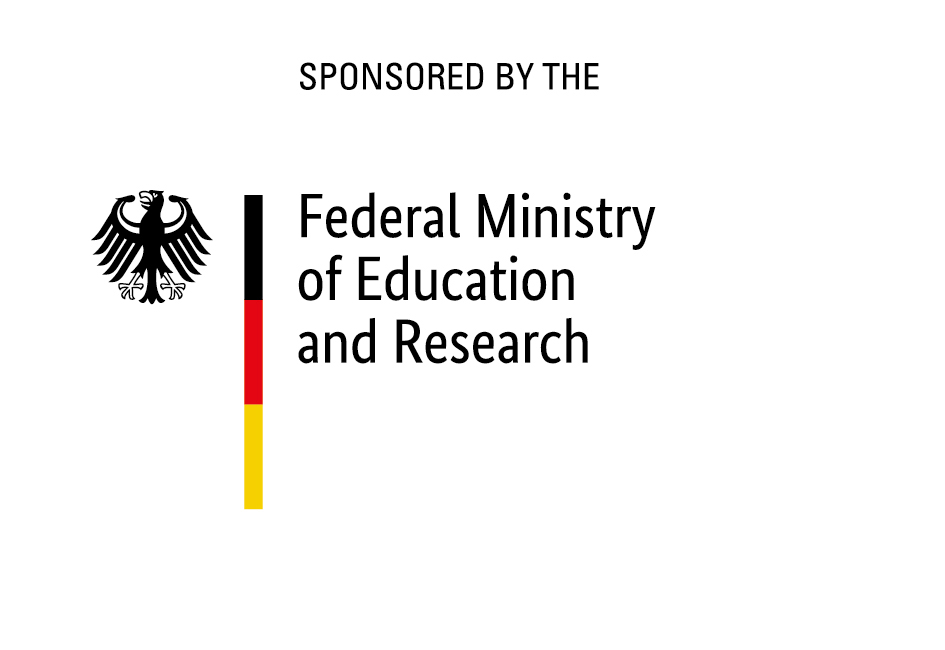}
    \label{fig:leximirka}
\end{figure}
The project on which this report is based was funded by the German Federal Ministry of Education and Research (BMBF) under the funding code 01|S20049. The author is responsible for the content of this publication.

\newpage

\section*{Ethical Statement}

In this paper, the authors introduce HateBERT, a pre-trained language model for the study of abusive language phenomena in social media in English. HateBERT is unique because (i.) it is based on further pre-training of an existing pre-trained language model (i.e., BERT \texttt{base-uncased}) rather than training it from scratch, thus reducing the environmental impact of its creation;~\footnote{The Nvidia V100 GPU we used is shared and it has a maximum number of continuous reserved time of 72 hours. In total, it took 18 days to complete the 2 million retraining steps.} (ii.) it uses a large collection of messages from communities that have been deemed to violate the content policy of a social media platform, namely Reddit, because of expressing harassment, bullying, incitement of violence, hate, offense, and abuse. The judgment on policy violation has been made by the community administrators and moderators. We consider this dataset for further pre-training more ecologically representative of the expressions of different abusive language phenomena in English than the use of manually annotated datasets. 

The collection of banned subreddits has been retrieved from a publicly available collection of Reddit, obtained through the Reddit API. From this collection, we generated the RAL-E dataset. RAL-E will be publicly released (it is accessible also at review phase in the Supplementary Materials). While its availability may have an important impact in boosting research on abusive language phenomena, especially by making natural interactions in online communities available, we are also aware of the risks of privacy violations for owners of the messages. This is one of the reasons why at this stage, we only make available in RAL-E the content of the message without metadata such as the screen name of the author and the community where the message was posted. 

There are numerous benefits from using such models to monitor the spread of abusive language phenomena in social media. Among them, we mention the following: (i.) reducing exposure to harmful content in social media; (ii.) contributing to the creation of healthier online interactions; and (iii.) promoting positive contagious behaviors and interactions~\cite{matias2019preventing}. Unfortunately, work in this area is not free from potentially negative impacts. The most direct is a risk of promoting misrepresentation. HateBERT is an intrinsically biased pre-trained language model. The fine-tuned models that can be obtained are not overgenerating the positive classes, but they suffer from the biases in the manually annotated data, especially for the offensive language detection task~\cite{sap-etal-2019-risk,davidson-etal-2019-racial}. Furthermore, we think that such tools must always be used under the supervision of humans. Current datasets are completely lacking the actual context of occurrence of a messsage and the associated meaning nuances that may accompany it, labelling the positive classes only on the basis of superficial linguistic cues. The deployment of models based on HateBERT ``in the wild'' without human supervision requires additional research and suitable datasets for training.

We see benefits in the use of HateBERT in research on abusive language  phenomena as well as in the availability of RAL-E. Researchers are encouraged to be aware of the intrinsic biased nature of HateBERT and of its impacts in real-world scenarios.

\newpage

\bibliographystyle{acl}
\bibliography{bibliography}

\newpage

\section*{Appendix A}
\captionsetup{labelformat=AppendixTables}
\setcounter{table}{0}

\begin{table}[!tbh]
    \centering
    \begin{tabular}{lr}
    \toprule
    \bf Subreddit & \bf Number of posts  \\ \midrule
apewrangling & 5 \\
beatingfaggots    &  3 \\
blackpeoplehate     &         16 \\
chicongo             &        15 \\
chimpmusic            &       35 \\
didntdonuffins        &       22 \\
fatpeoplehate     &      1465531 \\
funnyniggers       &          29 \\
gibsmedat          &          24 \\
hitler             &         297 \\
holocaust          &        4946 \\
kike               &           1 \\
klukluxklan        &           1 \\
milliondollarextreme   &    9543 \\
misogyny    &                390 \\
muhdick      &                15 \\
nazi         &              1103 \\
niggas        &               86 \\
niggerhistorymonth      &     28 \\
niggerrebooted          &      5 \\
niggerspics             &    449 \\
niggersstories          &     75 \\
niggervideos        &        311 \\
niglets             &         27 \\
pol                 &         80 \\
polacks             &        151 \\
sjwhate             &      10080 \\
teenapers           &         23 \\
whitesarecriminals      &     15 \\
    \end{tabular}
    \caption{Distribution of messages per banned community composing the RAL-E dataset.}
    \label{tab:rale}
\end{table}

\paragraph{Pre-processing before re-training}

\begin{itemize}
    \item all users' mentions have been substituted with a placeholder (@USER);
    \item all URLs have been substituted with a with a placeholder (URL);
    \item emojis have been replaced with text (e.g. \includegraphics[height=1em]{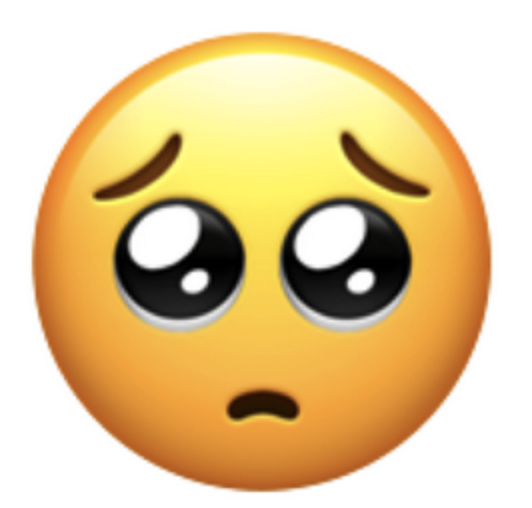} $\rightarrow$ \texttt{:pleading\_face:}) using Python \texttt{emoji} package;
    \item hashtag symbol has been removed from hasthtags (e.g. \#kadiricinadalet $\rightarrow$ kadiricinadalet);
    \item extra blank spaces have been replaced with a single space;
    \item extra blank new lines have been removed.
\end{itemize}

\section*{Appendix B}

\paragraph{Pre-processing before fine-tuning} For each dataset, we have adopted minimal pre-processing steps. In particular:

\begin{itemize}
    \item all users' mentions have been substituted with a placeholder (@USER);
    \item all URLs have been substituted with a with a placeholder (URL);
    \item emojis have been replaced with text (e.g. \includegraphics[height=1em]{emoji} $\rightarrow$ \texttt{:pleading\_face:}) using Python \texttt{emoji} package;
    \item hashtag symbol has been removed from hashtags (e.g. \#kadiricinadalet $\rightarrow$ kadiricinadalet);
    \item extra blank spaces have been replaced with a single space.
\end{itemize}

\begin{table}[!tbh]
    \centering
    \begin{tabular}{lr}
    \toprule
        \bf Hyperparameters &  \bf Value  
        \\ \midrule
         Learning rate & 1e-5 \\  
         Training Epoch & 5 \\
         Adam epsilon & 1e-8 \\
         Max sequence length & 100 \\
         Batch size & 32 \\
         Num. warmup steps & 0 \\
    \end{tabular}
    \caption{Hyperparamters for fine-tuning BERT and HateBERT.}
    \label{tab:hyper}
\end{table}


%

\end{document}